\documentclass[runningheads]{llncs}
\usepackage{graphicx}
\usepackage{amsmath,amssymb} 
\usepackage{color}

\usepackage{times}
\usepackage{epsfig}

\usepackage{url}
\usepackage[caption=false]{subfig}
\usepackage{mdwlist}
\usepackage{multirow}
\usepackage{cite}

\graphicspath{{figs/}}

\usepackage{bm}
\newcommand{\bb}[1]{\boldsymbol{\mathrm{#1}}}
\newcommand{\Tr}{\mathrm{T}}

\begin{document}
\title{ForestHash: Semantic Hashing With Shallow Random Forests and Tiny Convolutional Networks} 

\titlerunning{ForestHash}
%

\author{Qiang Qiu\inst{1} \and
Jos\'e Lezama\inst{2} \and
Alex Bronstein\inst{3} \and
Guillermo Sapiro\inst{1}}
%
\authorrunning{Q. Qiu, J. Lezama, A. Bronstein, and G. Sapiro}
%

\institute{ Duke University, USA \and
Universidad de la Rep\'ublica, Uruguay \and
Technion-Israel Institute of Technology, Israel}
\maketitle              
\begin{abstract}
In this paper, we introduce a random forest semantic hashing scheme that embeds
tiny convolutional neural networks (CNN) into shallow random forests.  A binary
hash code for a data point is obtained by a set of decision trees, setting `1'
for the visited tree leaf, and `0' for the rest.  We propose to first randomly
group arriving classes at each tree split node into two groups, obtaining a
significantly simplified two-class classification problem that can be a handled
with a light-weight CNN weak learner. Code uniqueness is achieved via the random
class grouping, whilst code consistency is achieved using a low-rank loss in the
CNN weak learners that encourages intra-class compactness for the two random class
groups. Finally, we introduce an information-theoretic approach for aggregating
codes of individual trees into a single hash code, producing a near-optimal
unique hash for each class. The proposed approach significantly outperforms
state-of-the-art hashing methods for image retrieval tasks on large-scale public
datasets, and is comparable to image classification methods while utilizing a
more compact, efficient and scalable representation. This work proposes a
principled and robust procedure to train and deploy in parallel an ensemble of
light-weight CNNs, instead of simply going deeper.
%
\end{abstract}
\section{Introduction}
\label{sec:intr}

\begin{figure*} [t]
\centering
 \includegraphics[angle=0, width=.8\textwidth]{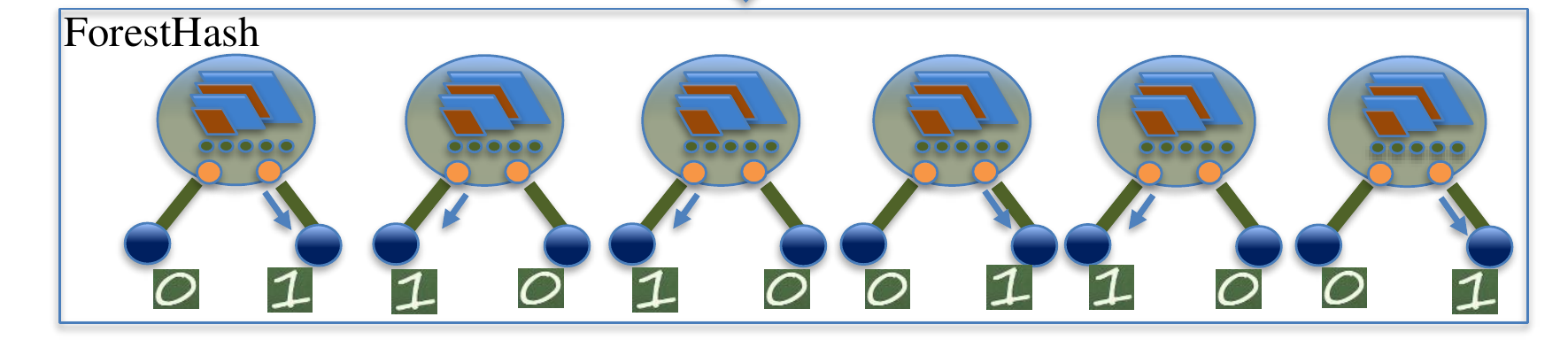}
\caption{ForestHash embeds tiny convolutional neural networks (CNN) into shallow
  random forests. ForestHash consists of shallow random trees in a forest,
  usually of depth 2 or 3. At each tree split node, arriving classes are
  randomly partitioned into two groups for a significantly simplified two-class
  classification problem, which can be sufficiently handled by a light-weight
  CNN weak learner, usually of 2 to 4 layers.  We set ‘1’ for the visited tree
  leaf, and ‘0’ for the rest. By simultaneously pushing each data point through
  $M$ trees of the depth $d$, we obtain $M (2^{d-1})$-bit hash codes.  The
  random grouping of the classes enables code uniqueness by enforcing that each
  class shares code with different classes in different trees. The
  non-conventional low-rank loss adopted for CNN weak learners encourages code
  consistency by minimizing intra-class variations and maximizing inter-class
  distance for the two random class groups.  The obtained ForestHash codes serve
  as efficient and compact image representation for both image retrieval and
  classification.  }
\label{fig:overview}
\end{figure*}

In view of the recent huge interest in image classification and object
recognition problems and the spectacular success of deep learning and random
forests in solving these tasks,  modest 
efforts are being invested into the related, and often more difficult, problems
of image and multimodal content-based retrieval, and, more generally, similarity
assessment in very large-scale databases. These problems, arising as primitives
in many computer vision tasks, are becoming increasingly important in the era of
exponentially increasing information.  Semantic and similarity-preserving
hashing methods have recently received considerable attention for addressing
such a need, in part due to their significant memory and computational advantage
over other representations.  These methods learn to embed data points into a
space of binary strings; thus producing compact representations with constant or
sub-linear search time; this is critical and one of the few options for low-cost
truly big data.  Such an embedding can be considered as a hashing function on
the data, which translates the underlying similarity into the collision
probability of the hash or, more generally, into the similarity of the codes
under the Hamming metric.  Examples of recent similarity-preserving hashing
methods include Locality-Sensitive Hashing (LSH) \cite{LSH} and its kernelized
version (KLSH) \cite{KLSH}, Spectral Hashing (SH) \cite{SH}, Sparse Hash
\cite{sparsehash}, Kernel-based Supervised Hashing (KSH) \cite{KSH}, Anchor
Graph Hashing (AGH) \cite{AGH}, Self-Taught Hashing (STH) \cite{STH}, and Deep
Supervised Hashing (DSH) \cite{DSH}.

Due to the profound similarity between the problems of semantic hashing and that
of binary classification, numerous classification techniques have been adapted
to the former task.  For example, multiple state-of-the-art supervised hashing
techniques like ANN Hashing \cite{MM-NN}, SparseHash \cite{sparsehash}, HDML
\cite{HDML} and DSH \cite{STH} are based on deep learning methodologies.
Besides deep learning, random forest \cite{RF2001, RFBook} is another popular
classification technique that has recently shown great success for a large
variety of classification tasks, such as pose estimation \cite{RF-pose} and
object recognition \cite{RF-obj}.  However, to the best of our knowledge, random
forests have not been used so far to construct semantic hashing schemes, and
therefore do not enjoy the advantages of such compact and efficient  codes.  This
is mainly because acting as hashing functions, a random forest fails to preserve
the underlying similarity due to the inconsistency of hash codes generated in
each tree for the same class data; it also lacks a principled way of aggregating
hash codes produced by individual trees into a single longer code.

In this paper, we propose the \emph{ForestHash} scheme.  As shown in Figure
\ref{fig:overview}, the proposed ForestHash is designed to provide consistent
and unique hashes to images from the same semantic class, by embedding tiny
convolutional neural networks (CNN) into shallow random forests.
  We start with a
simple hashing scheme, where random trees in a forest act as hashing functions
by setting `1' for the visited tree leaf, and `0' for the rest.  To enable such
hashing scheme, we first introduce random class grouping to randomly partition
arriving classes into two groups at each tree split node. The class random
grouping enables code uniqueness by enforcing each class to share code with
different classes in different trees, and also produces a significantly reduced
two-class problem being sufficiently handled by a light-weight CNN.

We further adopt a non-conventional low-rank loss for CNN weak learners to
encourage code consistency by minimizing intra-class variations and maximizing
inter-class distance for the two random class groups, thereby preserving
similarity.  The low-rank loss is based on the assumption that high-dimensional
data often have a small intrinsic dimension.  Consequently, when data from the
same low-dimensional subspace are arranged as columns of a single matrix, this
matrix should be approximately low-rank.  In Section \ref{sec:trans}, we show
how to learn a linear transformation of subspaces using the matrix nuclear norm
as the optimization criterion. We discuss both experimentally and theoretically
that such learned transformation simultaneously minimizes intra-class variation
and maximizes inter-class separation. We further show that kernelization or deep
learning can be used to handle intricate data that do not necessarily admit a
linear model.

Finally, the proposed information-theoretic aggregation scheme provides a
principled way to combine hashes from each independently trained random tree in
the forest. The aggregation process discussed in Section \ref{sec:it} is
performed efficiently in a greedy way, which still achieves a near-optimal
solution due to submodularity of the mutual information criterion being
optimized.  We discuss both unsupervised and supervised hash aggregation.

In Section~\ref{sec:exp}, we show a comprehensive experimental evaluation of the
proposed representation scheme, demonstrating that it significantly outperforms
state-of-the-art hashing methods for large-scale image and multi-modal retrieval
tasks.

\section{Forest hashing}
\label{sec:thm}

We first discuss a simple random forest hashing scheme, where independently
trained random trees act as hashing functions by setting `1' for the visited
tree leaf, and `0' for the rest.  We also show that hashes from a forest often
fail to preserve the desired intra-class similarity.

\subsection{A toy hashing scheme}

{Random forest} \cite{RF2001, RFBook} is an ensemble of binary \emph{decision
  trees}, where each tree consists of hierarchically connected \emph{split}
(internal) nodes and \emph{leaf} (terminal) nodes.  Each split node corresponds
to a \emph{weak learner}, and evaluates each arriving data point sending it to
the left or right child based on the weak learner binary outputs.  Each leaf
node stores the statistics of the data points that arrived to it during
training.  During testing, each decision tree returns a class posterior
probability for a test sample, and the forest output is often defined as the
average (or otherwise aggregated distribution) of the tree posteriors.

Following the random forest literature \cite{RFBook}, in this paper, we specify
a maximum tree depth $d$ to limit the size of a tree, which is different from
algorithms like C4.5 \cite{c4.5} that grow the tree relying on other termination
criteria; we also avoid post-training operations such as tree pruning. Thus, a
tree of depth $d$ consists of $2^{d-1}$ tree leaf nodes, indexed in the
breadth-first order.

During training, we can introduce randomness into the forest through a
combination of random set sampling and randomized node optimization, thereby
avoiding duplicate trees.  As discussed in \cite{RF2001, RFBook}, training each
tree with a different randomly selected set decreases the risk of overfitting,
improves the generalization of classification forests, and significantly reduces
the training time. When given more than two classes, we introduce node
randomness by \emph{randomly} partitioning the classes arriving at each binary
split node into two categories.

A toy pedagogic hashing scheme is constructed as follows: Each data point is
pushed through a tree until reaching the corresponding leaf node. We simply set
`1' for leaf nodes visited, and `0' for the rest.  By ordering those bits in a
predefined node order, e.g., the breadth-first order, we obtain a
$(2^{d-1})$-bit hash code, always containing exactly one 1.  In a random forest
consisting of $M$ trees of the depth $d$, each point is simultaneously pushed
through all trees to obtain $M$ $(2^{d-1})$-bit hash codes.

This hashing scheme has several obvious characteristics and advantages: First,
both the training and the hashing processes can be done in parallel to achieve
high computational efficiency on modern parallel CPU or GPU hardware.  Second,
multiple hash codes obtained from a forest, each from an independently trained
tree, have the potential to inherit the boosting effect of the random forest,
i.e., increasing the number of trees increases accuracy (sublinearly)
\cite{RFBook}.  Finally, the scheme guarantees $1$-sparsity for hash codes from
each tree.

However, hashes from a forest fail to preserve the underlying data similarity.
In classification, for which the forest was originally designed, an ensemble
posterior is obtained by averaging from a large number of trees, thus boosting
the classification accuracy \cite{RF2001}, and no confident class posteriors are
required for individual trees. This has several negative consequences for
constructing a suitable hash function. First, a forest often distributes same
class samples over multiple leave nodes in a tree, thus, no consistent codes are
assigned to each class. Second, for the same reason, samples of different
classes can follow the same path, therefore a forest does not guarantee a unique
code for each class. Moreover, it is not obvious how to combine hashes from
different trees given a target code length.

The inconsistency of the hash codes becomes more severe when increasing the tree
depth, as more leaf nodes are available to distribute the same class
samples. This problem can not be solved by simply increasing the number of trees
for longer total bit length. For example, if 4-bit inconsistency is allowed for
a 64-bit hash code, the Hamming ball already contains $ C_{64}^4= 635,376$
codes.  A principled way is required to combine hashes from each tree. One can
choose to combine hashes from different trees simply through concatenating,
averaging and thresholding, or voting. However, the principles behind those
heuristics are not obvious, and we might loose control on code length, sparsity,
and even binarity.

In what follows, we address these two problems. First, we propose the random class
grouping scheme, followed with near-optimal code aggregation, to enforce code
uniqueness for each class. Second, we adopt a non-conventional low-rank loss for
weak learners to encourage code consistency.

\subsection{Random class grouping} \label{sec:grouping}
A random class grouping scheme is first introduced to randomly partition
arriving classes into two groups at each tree split node. Random class grouping
serves two main purposes: First, a multi-class problem is significantly reduced
to a two-class classification problem at each split node, which can be
sufficiently handled by a very light-weight CNN weak learner.  Second, random class
grouping enforces each class to share its code with different classes in different
trees, which allows the information-theoretic aggregation developed in the
sequel to later produce a near-optimal unique hash code for each class.

\subsection{Low-rank loss}
\label{sec:trans}

\begin{figure*} [t]
  \centering
  \subfloat[] {\label{fig:x} \includegraphics[angle=0, width=.18\textwidth]{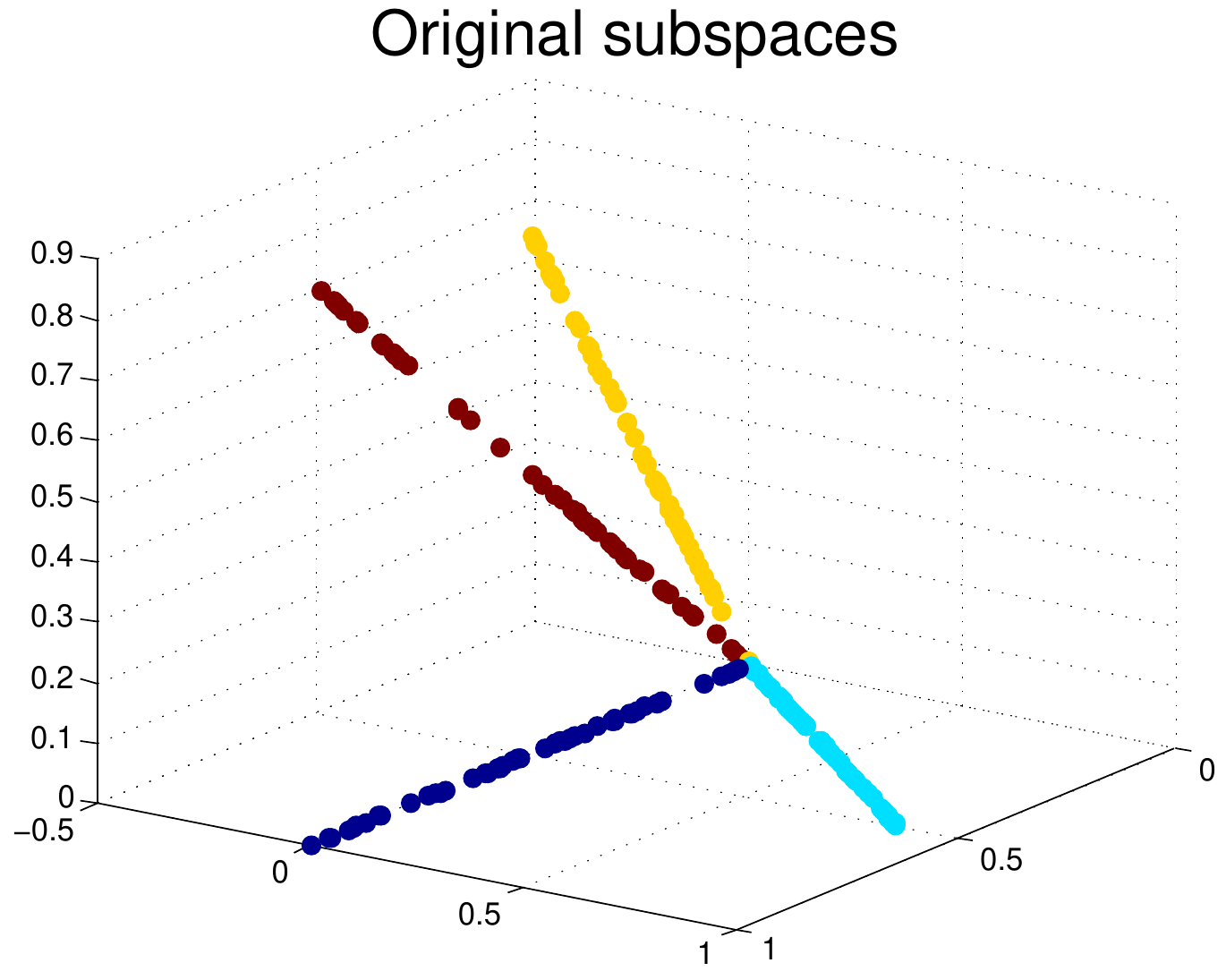} \hspace{-2pt}}
  \subfloat[] {\label{fig:x} \includegraphics[angle=0, width=.18\textwidth]{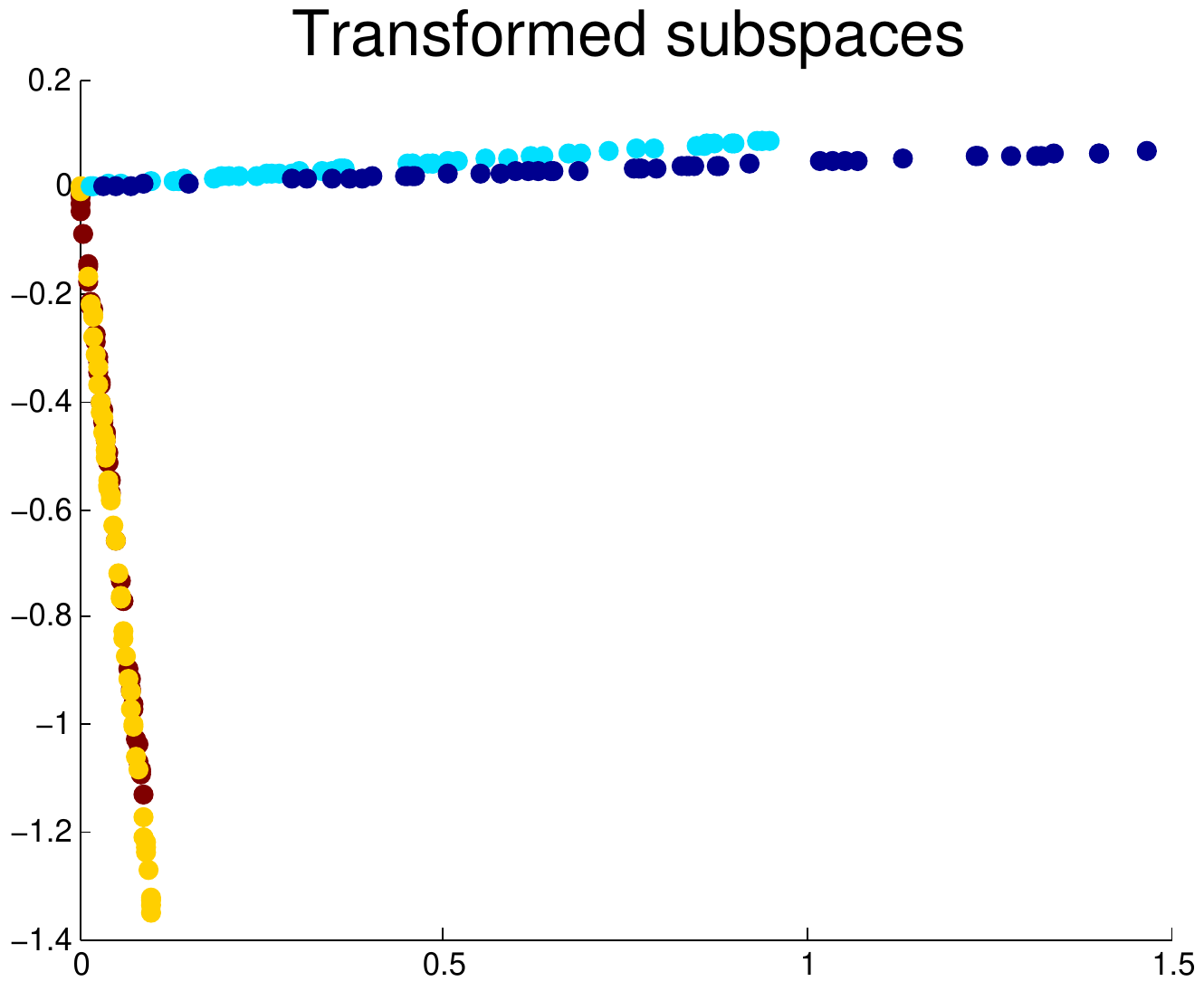} \hspace{8pt}}
  \subfloat[] {\label{fig:x} \includegraphics[angle=0,  width=.18\textwidth]{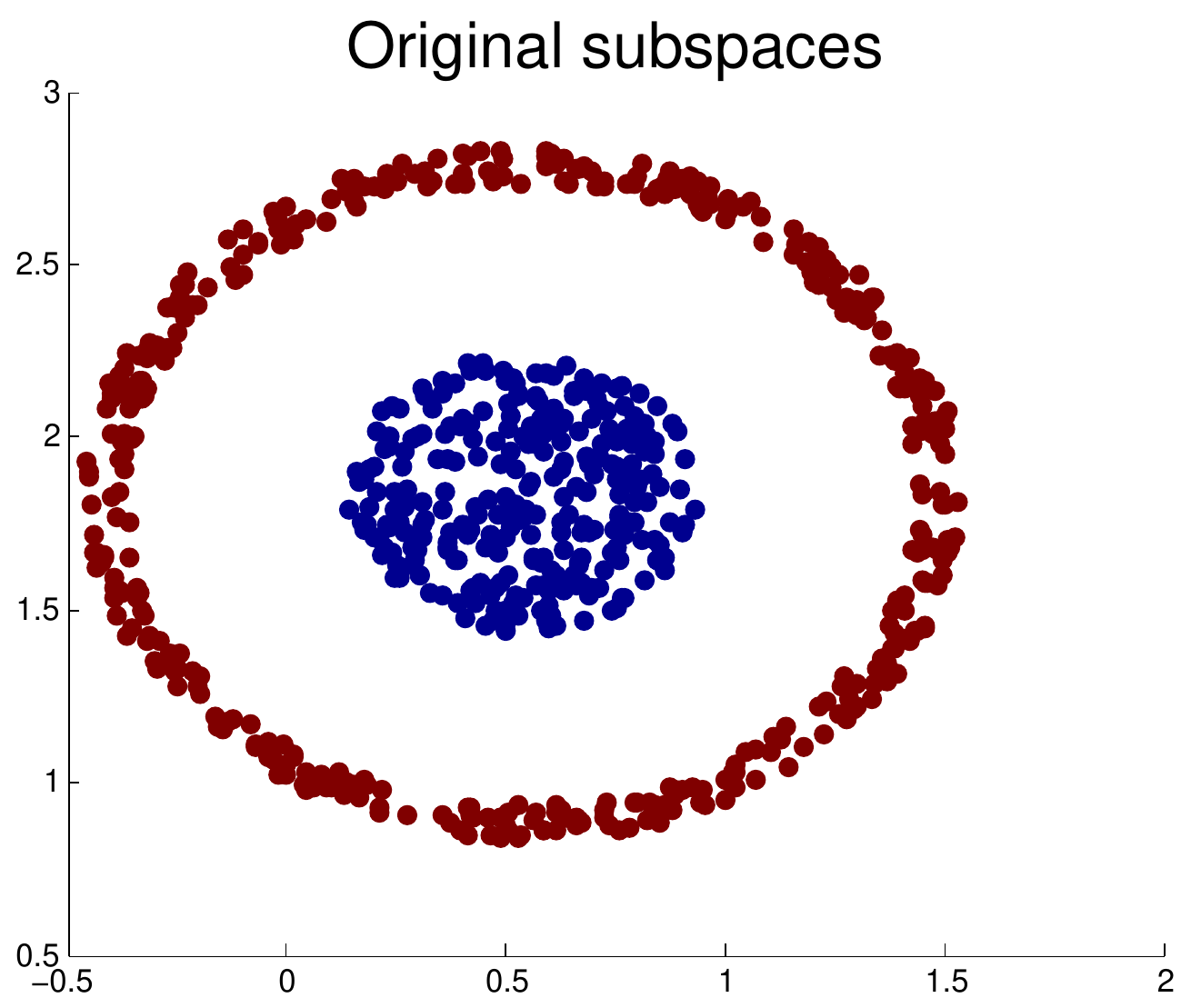} \hspace{-2pt}}
  \subfloat[] {\label{fig:x} \includegraphics[angle=0, width=.18\textwidth]{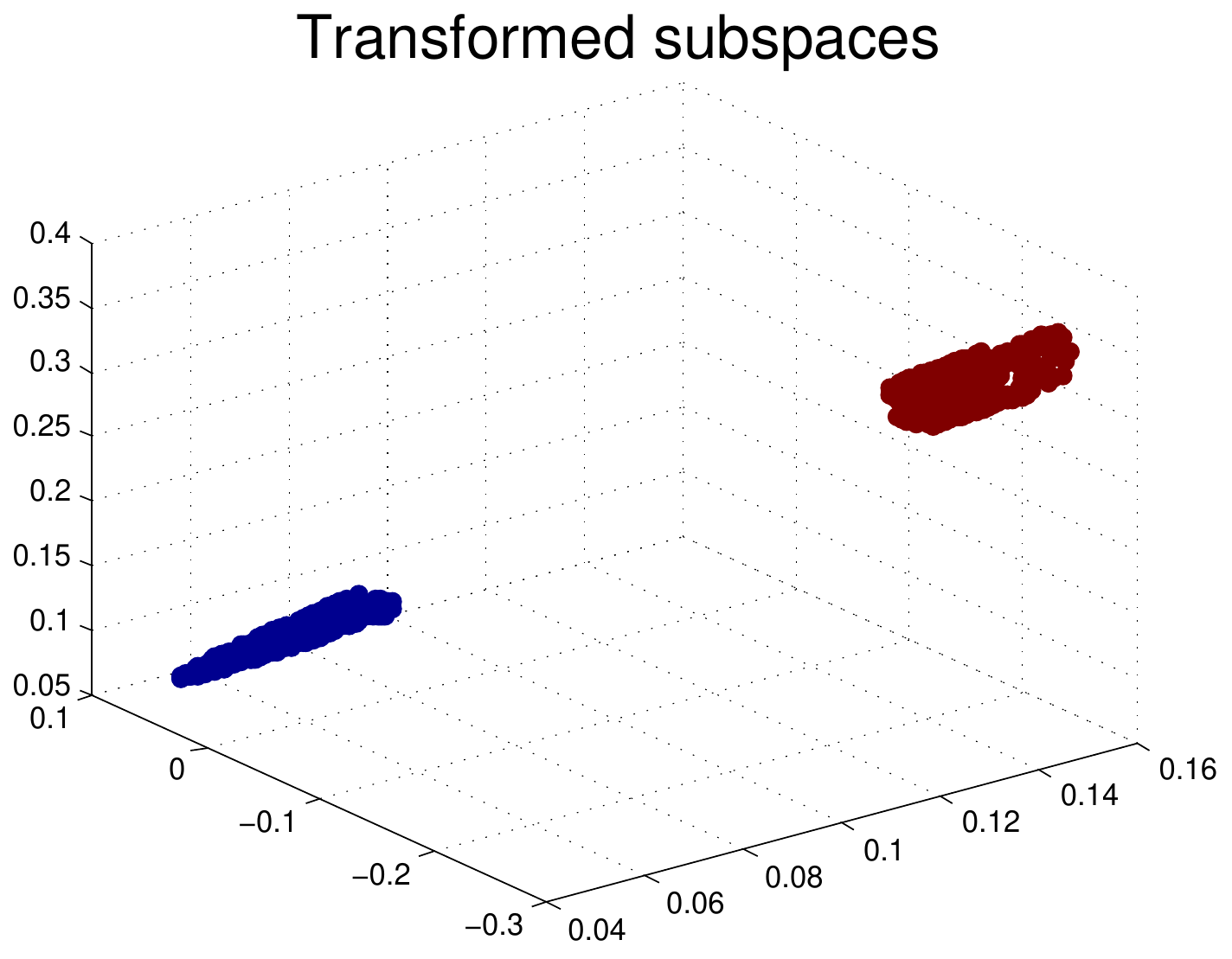} \hspace{8pt}}
  \caption{Synthetic two-class examples illustrating the properties of the
    learned low-rank transformation. (a), (c) are transformed to (b), (d),
    respectively. In (a), two classes are defined as \{blue, cyan\} and
    \{yellow, red\}. An RBF kernel is applied to transform (c) to (d)}
  \label{fig:toyexample}
\end{figure*}


A  non-conventional low-rank loss is adopted
for weak learners, e.g., a light-weight CNN learner, in a forest.  Consider $s$-dimensional data points belonging
to two classes after random class grouping, which for simplicity are denoted as
positive and negative. We stack the points as columns of the matrices
$\mathbf{X}^+$ and $\mathbf{X}^-$, respectively.  Let $||\mathbf{A}||_*$ denote
the \emph{nuclear norm} of the matrix $\mathbf{A}$, i.e., the sum of its
singular values. The nuclear norm is known to be the convex envelope of matrix
rank over the unit ball of matrices \cite{rank-min}. The following result in
\cite{lowrankT} helps motivate our per-node classifier:

\begin{lemma}  \label{nuclear_ineq}
Let $\mathbf{A}$ and $\mathbf{B}$ be matrices of the same row dimensions, and
$\mathbf{[A,B]}$ denote their column-wise concatenation. Then, $
||\mathbf{[\mathbf{A},\mathbf{B}]}||_* \le ||\mathbf{A}||_* + ||\mathbf{B}||_*,
$ with equality holding if the column spaces of $\mathbf{A}$ and $\mathbf{B}$
are orthogonal.
\end{lemma}

At each tree split node, we propose to learn a
weight matrix $\mathbf{W}$  minimizing
the following low-rank loss function.
\begin{align} \label{nuclear_obj} \nonumber
\underset{\mathbf{W}} \min ||\mathbf{W X}^+||_* + ||\mathbf{W X}^-||_* -& ||\mathbf{W [X^+, X^-]}||_*,\\ 
\end{align}
Based on Lemma~\ref{nuclear_ineq}, the loss function (\ref{nuclear_obj}) reaches
its minimum $0$ if the column spaces of the two classes become orthogonal after
applying the learned transformation $\mathbf{W}$. Equivalently,
(\ref{nuclear_obj}) reaches the minimum $0$ if the subspaces of the two classes
are maximally opened up after transformation, i.e., the smallest principal angle
between the subspaces equals $\frac{\pi}{2}$. Simultaneously, minimizing the
first two nuclear norm terms in (\ref{nuclear_obj}) helps reduce the variation
within classes.  Synthetic examples presented in Figure~\ref{fig:toyexample}
illustrate the properties of the learned transformation.  The trivial solution
$\mathbf{W}=0$ can be avoided through a good initialization, e.g., the identity matrix \cite{lowrankT}.

\textbf{Splitting functions.}  With random class grouping, we have a two-class
classification problem at each split node.  We stack the training data points
from each class as columns of the matrices $\mathbf{X}^+$ and $\mathbf{X}^-$,
respectively.

During training, at the $i$-th split node, we denote the arriving training
samples as $\mathbf{X}^+$ and $\mathbf{X}^-$.  After a weight matrix
$\mathbf{W}$ is successfully learned by minimizing (\ref{nuclear_obj}), it is
reasonable to assume that each of the classes will belong to a low-dimensional
subspace, the distance from which can be used to classify previously unseen
points.  We use $k$-SVD \cite{Elad_KSVD} to learn a pair of dictionaries
$\bb{D}^{\pm}$, for each of the two classes, by minimizing
\begin{eqnarray}
\min_{\bb{D}_\pm, \bb{Z}^\pm} \|  \bb{WX}^\pm - \bb{D}^\pm \bb{Z}^\pm \|  & \mathrm{s.t.} & \| \bb{z}^\pm \|_0 \le l,
\end{eqnarray}
where the $\ell_0$ pseudonorm $\| \bb{z}^\pm \|_0$ counts the number of non-zero elements in each column of $\bb{Z}^\pm$, and $l$ controls the subspace dimension.

At testing, given a data point $\bb{x}$, the splitting function is evaluated by first projecting $\bb{x}$ onto both dictionaries and evaluating the projection errors
\begin{equation}
e^\pm(\bb{x}) = \mathrm{arg}\min_{\bb{z}^\pm} \| \bb{D}^\pm \bb{z}^\pm - \bb{Wx} \|_2 \, =\,  \| \bb{P}^\pm \bb{x} \|_2,
\end{equation}
where $\bb{P}^\pm = \bb{D}^\pm (\bb{D}^{\pm \Tr} \bb{D}^\pm)^{-1} \bb{D}^{\pm \Tr} \bb{W}$ are the $n \times n$ projection matrices. The point is sent to the left subtree if $e^-(\bb{x}) < e^+(\bb{x})$, and to the right subtree otherwise. In practice, we only store the projection matrices $\bb{P}^\pm$ at each split node.
Note that similar splitting functions report success in a classification context with much deeper trees in \cite{qiu2013learning}.

\textbf{Optimization.}  To optimize the low-rank loss function
(\ref{nuclear_obj}) using gradient descent, the subgradient of the nuclear norm
of a matrix can be computed as follows: Let $\mathbf{A}=\mathbf{U \Sigma
  V}^\mathrm{T}$ be the SVD decomposition of the matrix A. Let
$\mathbf{\hat{U}}$ and $\mathbf{\hat{V}}$ be the columns of $\mathbf{U}$ and
$\mathbf{V}$ corresponding to eigenvalues larger than a predefined threshold.
Following \cite{lowrankT,subdifferential}, the subgradient of the nuclear norm
can be evaluated in a simplified form as
\[
\partial||\mathbf{A}||_* = \mathbf{\hat{U}} \mathbf{\hat{V}} ^\mathrm{T} 
\]
Note that (\ref{nuclear_obj}) is a D.C. (difference of convex functions)
program; and the minimization is guaranteed to converge to a local minimum (or a
stationary point), with the D.C. procedure detailed in \cite{dc2,dc1}.

\textbf{Kernelization.}  A sufficient number of tree splits could potentially
handle non-linearity in data for classification. In this work, only very limited
number of splits is preferred in each tree, e.g., depth 1 to 3, to encourage
short codes, which is insufficient in modeling data non-linearity
well. Moreover, if we rely on tree splits in modeling non-linearity, we may
still obtain less confident class posteriors as explained.  The low-rank loss in
(\ref{nuclear_obj}) is particularly effective when data approximately lie in
linear subspaces \cite{lowrankT}.  To improve the ability of handling more
generic data, an effective way is to map data points into an inner product space
prior to optimize for low-rank loss.

Given a data point $\mathbf{y}$, we create a nonlinear map
$\mathcal{K}(\mathbf{x}) = (\kappa(\mathbf{x}, \mathbf{x}_1); ...;
\kappa(\mathbf{x}, \mathbf{x}_n))$ by computing the inner product between
$\mathbf{x}$ and a fixed set of $n$ points $\{\mathbf{x}_1, ..., \mathbf{x}_n\}$
randomly drawn from the training set. The inner products are computed via the
kernel function, $\kappa(\mathbf{x}, \mathbf{x}_i) = \varphi(\mathbf{x})'
\varphi(\mathbf{x}_i)$, which has to satisfy the Mercer conditions; note that no
explicit representation for $\varphi$ is required. Examples of kernel functions
include polynomial kernels $\kappa(\mathbf{y}, \mathbf{x}_i) =
(\mathbf{x}'\mathbf{x}_i+p)^q$ (with $p$ and $q$ being constants), and radial
basis function (RBF) kernels $\kappa(\mathbf{x}, \mathbf{x}_i) = \exp
(-\frac{||\mathbf{x}-\mathbf{x}_i||_2^2}{2\sigma^2})$ with variance $\sigma^2$.
Given the data points $\mathbf{X}$, the set of mapped data is denoted as
$\mathcal{K}(\mathbf{X}) \subseteq \mathbb{R}^n$.  We now learn a weight matrix
$\mathbf{W}$ minimizing,
\begin{align} \label{knuclear_obj} \nonumber
\underset{\mathbf{W}}  \min ||\mathbf{W} \mathcal{K}(\mathbf{X}^+)||_* + ||\mathbf{W} \mathcal{K}(\mathbf{X}^-)||_* -& ||\mathbf{W} [\mathcal{K}(\mathbf{X}^+), \mathcal{K}(\mathbf{X}^-)]||_*, \\
\end{align}

\begin{figure} [t!]
  \begin{center}
\begin{tabular}{cc} 
 \includegraphics[width=.27\textwidth]{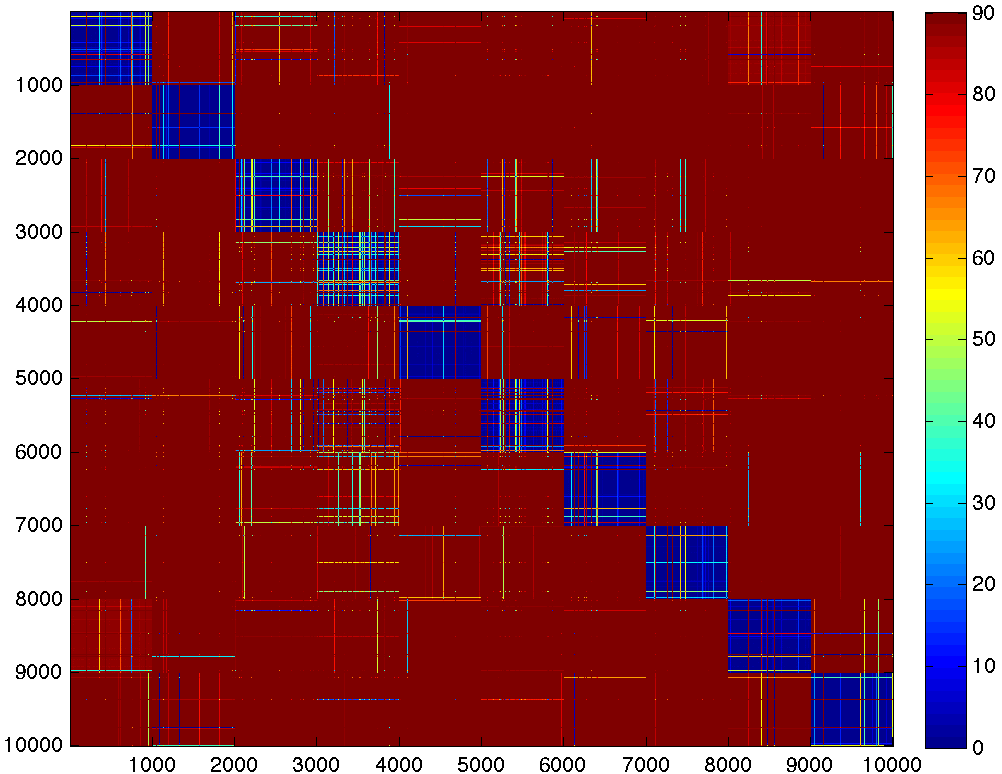} & %
 \includegraphics[width=.27\textwidth]{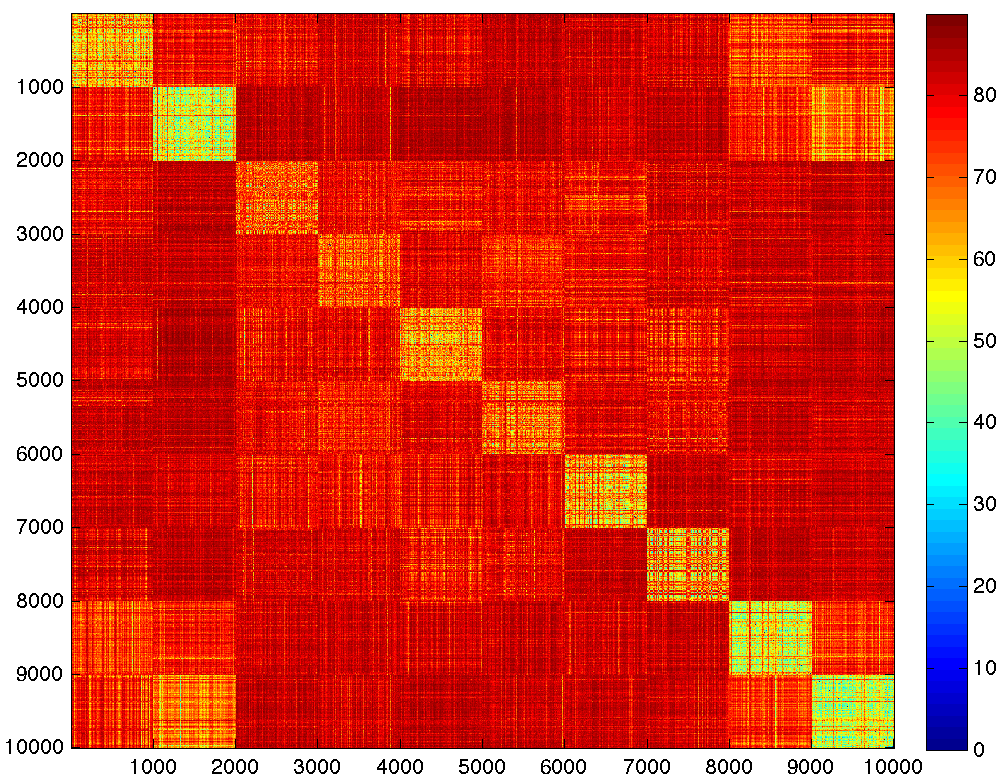}\\ %
\end{tabular} 
\end{center} 
  \caption{
  Angles between the deep features learned for the
    validation set of CIFAR-10 using VGG-16.  (Left) with additional low-rank
    loss. (Right) with the standard softmax loss.  With low-rank loss, the
    intra-class variations among features are collapsed and inter-class features
    are orthogonal, which are particularly preferred at each tree split
    node.}
\label{fig:cifar_tsne}
\end{figure}

\textbf{Deep networks.}  While kernelization shows a simple yet effective
non-linear mapping, we present a CNN-based weak learner now as the ultimate way in handling intricate
data.  With the gradient descent optimization discussed above, it is
possible to implement the following function
\begin{align} \label{lowrank-loss} \nonumber
L = ||\mathbf{\Phi}(\mathbf{X}^+)||_* + ||\mathbf{\Phi}(\mathbf{X}^-)||_* -& ||[\mathbf{\Phi}(\mathbf{X}^+), \mathbf{\Phi}(\mathbf{X}^-)]||_*, \\
\end{align}
as a low-rank loss layer for general deep networks, where $\mathbf{\Phi}$
denotes the mapping from a deep network.  From our
experimental experience, the low-rank loss reports comparable performance as the
standard softmax loss, while being used standalone as a classification loss for
small classification problems. However, together with softmax, we observed
consistent classification performance improvements over most popular CNN
architectures and challenging datasets. As in Fig.~\ref{fig:cifar_tsne}, with low-rank
loss, the intra-class variations among features are collapsed and inter-class
features are orthogonal \cite{Lezama2018OLE}. Such property is particularly beneficial at each tree
split node.

\subsection{Information-theoretic code aggregation}
\label{sec:it}

After training each random tree with the low-rank loss learner to produce consistent
hashes for similar data points, we propose an information-theoretic approach to
aggregate hashes across trees into a unique code for each data class. As labels
are usually unavailable or only available for a small subset of data,
unsupervised aggregation allows exploiting all available data. We also explain
how labels, if available, can be further incorporated for supervised hash
aggregation.  Note that the code aggregation step is only learned once during
training, no cost at testing.

\textbf{Unsupervised aggregation.} Consider a random forest consisting of $M$
trees of depth $d$; the hash codes obtained for $N$ training samples are denoted
as $\mathcal{B} = \{ \mathbf{B}_i\}_{i=1}^M$, with the $\mathbf{B}_i \in
\{0,1\}^{(2^{d-1}) \times N}$ being the codes generated from the $i$-th tree,
henceforth denoted as \emph{code blocks}.  Given the target hash code length
$L$, our objective is to select $k$ code blocks $\mathbf{B}^*$, $k \le
L/(2^{d-1})$, maximizing the mutual information between the selected and the
remaining codes,
\begin{equation}
\mathbf{B}^* = \arg \max_{\mathbf{B}: |\mathbf{B}|=k } I(\mathbf{B}; \mathcal{B} \backslash \mathbf{B}).
 \label{eqt:infomax1}
\end{equation}
A set function is said to be \emph{submodular} if it has a diminishing return
property, i.e., adding an element to a smaller set helps more than adding it to
a larger set.
\begin{lemma}\label{thm:submodular}
$I(\mathbf{B}; \mathcal{B} \backslash \mathbf{B})$ is submodular.
\end{lemma}

The general problem of maximizing submodular functions is NP-hard, by reduction
from the max-cover problem.  However, motivated by the sensor placement strategy in \cite{Guestrin08},
we propose a very simple greedy algorithm to approximate the
solution of (\ref{eqt:infomax1}). We start with $B=\emptyset$, and iteratively
choose the next best code block $\mathbf{b}^*$ from $\mathcal{B} \backslash
\mathbf{B}$ which provides a maximum increase in mutual information, i.e.,
\begin{eqnarray} \nonumber
\arg \max_{b^* \in \mathcal{B} \backslash \mathbf{B}}  I(\mathbf{B} \cup \mathbf{b}^*; \mathcal{B}  \backslash (\mathbf{B} \cup \mathbf{b}^*)) - I(\mathbf{B}; \mathcal{B} \backslash \mathbf{B}) \\
 = \arg \max_{b^* \in \mathcal{B} \backslash \mathbf{B}} H(\mathbf{b}^* | \mathbf{B}) - H(\mathbf{b}^* | \mathcal{B}  \backslash (\mathbf{B} \cup \mathbf{b}^*)),
 \label{eqt:greedy1}
\end{eqnarray}
 where $H(\mathbf{b}^* | \mathbf{B})$ denotes the conditional entropy.
 Intuitively, the first term $H(\mathbf{b}^* | \mathbf{B})$ forces
 $\mathbf{b}^*$ to be most different from the already selected codes
 $\mathbf{B}$, and the second term $- H(\mathbf{b}^* | \mathcal{B} \backslash
 (\mathbf{B} \cup \mathbf{b}^*))$ forces $\mathbf{b}^*$ to be most
 representative among the remaining codes. By defining a covariance matrix with
 the $ij$-th entry equal to $\exp(-\frac{d_{\mathbb{H}}(\mathbf{B}_i,
   \mathbf{B}_j)}{N})$, with $d_{\mathbb{H}}$ being the Hamming distance,
 (\ref{eqt:greedy1}) can be efficiently evaluated in a closed form as detailed
 in \cite{Guestrin08}.
 It has been proved in \cite{Guestrin08, submodular-bound}
  that the above greedy algorithm gives a polynomial-time
 approximation that is within $(1-1/e)$ of the optimum, where $e$ is the the
 Napier's constant.  Based on similar arguments as those in~\cite{Guestrin08},
 the near-optimality of our approach can be guaranteed if the forest size
 $|\mathcal{B}|$ is sufficiently larger than $2k$.

\textbf{Supervised aggregation.} When the class labels ${C}$ are available for
the $N$ training samples, an upper bound on the Bayes error over hashing codes
$\mathbf{B}$ is given by $\frac{1}{2}(H({C})-I(\mathbf{B}; {C}))$
\cite{Hellman70}.  This bound is minimized when $I(\mathbf{B}; {C})$ is
maximized. Thus, discriminative hash codes can be obtained by maximizing
\begin{equation}
\arg \max_{\mathbf{B}: |\mathbf{B}|=k} I(\mathbf{B}; {C}).
 \label{eqt:infomax2}
\end{equation}
Similarly to the unsupervised case, we maximize (\ref{eqt:infomax2}) using a
greedy algorithm initialized with $\mathbf{B}=\emptyset$ and iteratively
choosing the next best code block $\mathbf{b}^*$ from $\mathcal{B} \backslash
\mathbf{B}$ which provides a maximum mutual information increase, i.e.,
\begin{equation}
\arg \max_{\mathbf{b}^* \in \mathcal{B} \backslash \mathbf{B}} I(\mathbf{B} \cup \mathbf{b}^*; {C}) - I(\mathbf{B}; {C}),
 \label{eqt:greedy2}
\end{equation}
where $I(\mathbf{B}; {C})$ is evaluated as $I(\mathbf{B}; {C}) = H(\mathbf{B}) -
\sum_{c=1}^{p} p(c)H(\mathbf{B} |c).$ Entropy measures here involve computation
of probability density functions $p(\mathbf{B})$ and $p(\mathbf{B}|c)$, which
can both be efficiently computed by counting the frequency of unique codes in
$\mathbf{B}$. Note that the number of unique codes is usually very small due to
the learned transformation step.

\textbf{Semi-supervised aggregation.} The above two aggregation models can be simply unified as
\begin{eqnarray} \nonumber
\arg \max_{b^* \in \mathcal{B} \backslash \mathbf{B}}   [ I(\mathbf{B} \cup \mathbf{b}^*; \mathcal{B}  \backslash (\mathbf{B} \cup \mathbf{b}^*)) - I(\mathbf{B}; \mathcal{B} \backslash \mathbf{B})] \\
 + \lambda [I(\mathbf{B} \cup \mathbf{b}^*; {C}) - I(\mathbf{B}; {C})].
 \label{eqt:goverall}
\end{eqnarray}
The two terms here can be evaluated using different samples to exploit all
labeled and unlabeled data. The parameter $\lambda$ in (\ref{eqt:goverall}) is
suggested to be estimated as the ratio between the maximal information gained
from a code block to each respective criteria, i.e., $\lambda = \frac{\max_i
  I(\mathbf{B}_i;\mathcal{B} \backslash \mathbf{B}_i) }{\max_i I(\mathbf{B}_i;
  C)}.$ Exploiting the diminishing return property, only the first greedily
selected code block based on (\ref{eqt:greedy1}) and (\ref{eqt:greedy2}) need to
be evaluated, which leads to an efficient process for finding $\lambda$.
Selecting using only semantic information gives a hash model that is less
robust, e.g., overfits to training data, than a model also concerning the actual
code representation.  As shown in the experiments, both unsupervised and
supervised aggregation approaches promote unique codes for each class, with
further improvements when both are unified.

\subsection{Multimodal hashing}
We can further extend ForestHash as a multimodal similarity learning approach.  It
is often challenging to enable similarity assessment across modalities, for
example, searching a corpus consisting of audio, video, and text using queries
from one of the modalities.  The ForestHash framework can be easily extended for
hashing data from multiple modalities into a single space.

At training, when multimodal data arrives at a tree split node, we simply
enforce the same random class partition for all modalities, and learn for each
modality a dictionary pair independently using the shared class partition.
During training, only the splitting function of one dominant (usually most
discriminant) modality is evaluated for each arriving data point; during
testing, based on the modality of an arriving point, the corresponding splitting
function acts independently.  As shown in Section~\ref{sec:exp}, ForestHash
significantly outperforms state-of-the-art hashing approaches on cross-modality
multimedia retrieval tasks.

\section{Experimental evaluation}
\label{sec:exp}

We present an experimental evaluation of ForestHash on image retrieval tasks
using standard hashing benchmarks: the {CIFAR-10} image dataset \cite{cifar10},
the {MNIST} image dataset \cite{MNIST}, and the Wikipedia image and document
dataset \cite{wikixmod}.  {CIFAR-10} is a challenging dataset of 60,000
$32\times32$ labeled color images with 10 different object categories, and each
class contains 6,000 samples.  {MNIST} consists of 8-bit grayscale handwritten
digit images of ``0" to ``9" with 7,000 examples per class, and a total of 70,000
images.

\begin{table}
  \begin{center}
    \scriptsize
    \setlength{\tabcolsep}{4pt}
    \begin{tabular}{l cc}
      & CNN2 & \\
      \hline
      1 & Conv+ReLU+MaxPool & $5 \times 5 \times 3 \times 64$ \\           
      2 & Conv+ReLU+MaxPool & $5 \times 5 \times 64 \times 32$ \\           
      3 & FC & output: 256 \\      
      \hline    \\ 
      & CNN4 & \\
      \hline    
      1 & Conv+ReLU+MaxPool & $5 \times 5 \times 3 \times 64$ \\           
      2 & Conv+ReLU+MaxPool & $5 \times 5 \times 64 \times 64$ \\           
      3 & Conv+ReLU+MaxPool & $5 \times 5 \times 64 \times 64$ \\           
      4 & Conv+ReLU+MaxPool & $5 \times 5 \times 64 \times 64$ \\           				  				  
      5 & FC & output: 256 \\                                            				       				  
      \hline               
    \end{tabular}
  \end{center}
  \caption{Network structures of light-weight CNN learners.
  }
  \label{tab:cnn}
  \vspace{-1em}
\end{table}  


\begin{table}
  \begin{center}
    \scriptsize
    \setlength{\tabcolsep}{4pt}
    \begin{tabular}{l|cccc}
      \hline
      Method             & 12-bit    & 24-bit      & 36-bit    & 48-bit     \\ 
      \hline 
      \hline 
      LSH \cite{LSH}     & 0.13      & 0.14        & 0.14      & 0.15       \\
      SH \cite{SH}       & 0.13      & 0.13        & 0.14      & 0.13       \\
      ITQ \cite{ITQ}     & 0.11      & 0.11        & 0.11      & 0.12       \\
      CCA-ITQ \cite{ITQ} & 0.17      & 0.20        & 0.21      & 0.22       \\  
      MLH \cite{MLH}     & 0.18      & 0.20        & 0.21      & 0.21       \\  
      BRE \cite{BRE}     & 0.16      & 0.16        & 0.17      & 0.17       \\  
      KSH \cite{KSH}     & 0.29      & 0.37        & 0.40      & 0.42       \\  
      CNNH \cite{CNNH}   & 0.54      & 0.56        & 0.56      & 0.56       \\  
      DLBHC \cite{DLBHC} & 0.55      & 0.58        & 0.58      & 0.59       \\  
      DNNH \cite{DNNH}   & 0.57      & 0.59        & 0.59      & 0.59       \\  
      DSH \cite{DSH}     & 0.62      & 0.65        & 0.66      & 0.68       \\ 
      \hline 
      ForestHash-CNN2    & 0.61      & 0.75        & 0.78      & 0.80       \\ 
      ForestHash-CNN4    & 0.70      & 0.80        & 0.82      & 0.84       \\
      ForestHash-VGG16   & \bf{0.76} & \bf{0.82}   & \bf{0.86} & \bf{0.89}  \\        
      \hline               
    \end{tabular}
  \end{center}
  \caption{Retrieval performance (mAP) of different hashing methods on
    CIFAR-10. All methods use the 32x32 RGB images as input. }
  \label{tab:CIFAR-hash-50k}
  \vspace{-1em}
\end{table}  
      
      
\begin{table}
  \begin{center}
    \scriptsize
    \setlength{\tabcolsep}{4pt}
    \begin{tabular}{l|cccc}
      \hline
      Method                                                & depth & params & CIFAR-10 \\ 
      \hline 
      \hline
      Network in Network \cite{Lin2014}                     & -     & -      & 10.41 \\
      All-CNN \cite{All-CNN} 			            & -     & -      & 9.08  \\
      Deeply Supervised Net \cite{Lee2015}                  & -     & -      & 9.69  \\                
      FractalNet \cite{larsson2017fractalnet}               & 21    & 38.6M  & 10.18 \\ 
      ResNet ( \cite{resnet-eccv16})                        & 110   & 1.7M   & 13.63 \\
      ResNet with Stochastic Depth \cite{resnet-eccv16}     & 110   & 1.7M   & 11.66 \\ 
      \hline 
      \multirow{2}{*}{ResNet (pre-activation)\cite{He2016}} & 164   & 1.7M   & 11.26 \\
      & 1001  & 10.2M  & 10.56 \\  
      \hline 
      ForestHash CNN2 128-bit  & 2 ($\times$ 64)      & 0.58M ($\times$ 64)  & 20.3  \\ 
      ForestHash CNN4 128-bit  & 4 ($\times$ 64)      & 0.38M ($\times$ 64)  & 16.47 \\ 
      ForestHash VGG16 128-bit & 16 ($\times$ 64)     & 20.1M ($\times$ 64)  & 11.03 \\                 
      \hline                                                                 
    \end{tabular}
  \end{center}
  \caption{ Error rates (\%) on CIFAR-10 image classification
    benchmark.  ForestHash performs at the level of other
    state-of-the-art image classification techniques while utilizing
    a very compact 128-bit only representation. }
  \label{tab:CIFAR-acc}
  \vspace{-1em}
\end{table}


\begin{table}[ht]
  \centering
  
  {\scriptsize
    \begin{tabular}{l|ll||ll}
      \hline
      & \multicolumn{2}{c||}{radius = 0}  & \multicolumn{2}{c}{radius $\le$ 2}  \\
      \hline
      Method &  Precision & Recall & Precision& Recall \\
      \hline
      \hline
      SH \cite{SH}                 & 5.90           & 0.01           & 21.00          &  0.25 \\
      KSH \cite{KSH}               & 8.50           & 0.07           & 21.41          &  0.66 \\
      AGH1 \cite{AGH}              & 29.48          & 0.21           & 30.55          &  0.41 \\
      AGH2 \cite{AGH}              & 29.92          & 0.24           & 30.13          &  0.58 \\
      SparseHash \cite{sparsehash} & 16.65          & 0.05           & 32.69          &  1.81 \\
      \hline
      {ForestHash (rand)}          & 31.37          & 2.74           & 32.25          &  4.90 \\ 
      {ForestHash (unsup)}         & 34.02          & 3.65           & 34.55          &  6.40 \\ 
      {ForestHash (sup)}           & 33.86          & 3.33           & 34.02          &  5.21 \\ 
      {ForestHash (semi)}          & \textbf{34.05} & \textbf{4.12}  & 33.73          &  \textbf{7.29} \\ 
      \hline
      \hline
      {ForestHash CNN4-softmax}    & 22.72          & 0.33           & 34.27          & 1.52 \\ 
      {ForestHash CNN2-softmax}    & 23.00          & 0.42           & 32.13          & 1.56 \\ 
      {ForestHash CNN4}            &  28.66         & 0.86           & \textbf{38.60} & 2.88 \\ 
      {ForestHash CNN2}            &  29.30         & 1.78           & 38.29          & 4.68 \\  
      \hline
    \end{tabular}
  }	
  \caption{Retrieval performance ($\%$) of different hashing methods (48-bit
    codes) on CIFAR-10 using reduced training. The methods on the top two groups
    use GIST features. For reference, the bottom group shows the
    performance of ForestHash with CNN features extracted from the 32x32 RGB
    images.  }
  \label{tab:cifar10}
\end{table}
            
      
\begin{table*}[ht]
  \centering
      {\scriptsize
      	\begin{tabular}{l|l|l l l|l l l|l l l}
      	  \hline
          & & \multicolumn{3}{c|}{6,000 samples per class}  & \multicolumn{3}{c|}{ 100 samples per class} & \multicolumn{3}{c}{ 30 samples per class} \\
       	  \cline{3-11}
          &Test &Train  &  &  & Train & &  & Train & &  \\ 	
          &time ($\mu$s) &time (s) & Prec. & Rec.  & time (s) & Prec. & Rec.  & time (s) & Prec. & Rec. \\
      	  \hline
      	  \hline
          HDML \cite{HDML}          & 10  & 93780  & 92.94          & 60.44          & 1505  & 62.52           & 2.19           & 458  & 24.28          & 0.21  \\
          FastHash \cite{fasthash}  & 115 & 865    & 84.70          & 76.60          & 213   & 73.32           & 33.04          & 151  & 57.08          & 11.77 \\
          TSH  \cite{tsh}           & 411 & 164325 & 86.30          & 3.17           & 21.08 & 74.00           & 5.19           & 2.83 & 56.86          & 3.94  \\
          \hline                                                    
          ForestHash                & 57  & 24.20  & 86.53          & 46.30          & 4.19  & 84.98           & 45.00          & 1.43 & 79.38          & 42.27 \\
          ForestHash CNN2           & 13  & 81.6   & \textbf{97.99} & \textbf{95.99} & 7     & \textbf{94.24}  & \textbf{74.02} & 2.69 & \textbf{89.56} & \textbf{46.36} \\
          \hline
      	  \hline
      	\end{tabular}
      }	
      \caption{36-bit retrieval performance (\%) on MNIST (rejection hamming
        radius 0) using different training set sizes. Test time is the average
        binary encoding time in microseconds ($\mathrm{\mu s}$).}
      \label{tab:mnist}
\end{table*}


As discussed in Section~\ref{sec:trans}, a low-rank weak learner at each tree
split node is allowed in various implementations. Without particular
specification, a 256-dimensional RBF kernelization is assumed.  We use the CNN
suffix when using a light-weight CNN as weak learner. Table~\ref{tab:cnn} shows
two network structures of light-weight CNN learners, {CNN2} and {CNN4}, adopted
in experiments. Unless otherwise specified, 128 trees are trained and
semi-supervised aggregation are used (with only training data).

Note that a shallow tree is preferred; and a deeper tree ($d \ge 8$) becomes
less preferred for (fast) retrieval, and loses the robustness gained from
randomness. A tree of depth 2 is assumed by default in this section.  In
practice, the choice of tree depth also depends of the target code length and
the level of parallelism supported, as each hash tree can be trained and
deployed independently in parallel.

\begin{table*}[ht]
  \centering
  
  {\scriptsize
    \begin{tabular}{l|l|l|l|l|l|l|l|l}
      \hline
      & \begin{tabular}{l}ForestHash \\CNN2 \end{tabular} & ForestHash &  USPLH \cite{USPLH} & SH \cite{SH} & KLSH \cite{KLSH} & SIKH \cite{SIKH} & AGH1 \cite{AGH} & AGH2 \cite{AGH} \\
      \hline
      \hline
      24 bits & \textbf{99.63 } & 82.99 & 46.99 & 26.99 & 25.55 & 19.47 & 49.97 & 67.38 \\
      \hline
      48 bits & \textbf{99.68 } & 86.09 & 49.30 & 24.53 & 30.49 & 19.72 & 39.71 & 64.10 \\
      \hline
    \end{tabular}
  }	
  \caption{Mean average precision (mAP  in \%) in percent of Hamming ranking  on MNIST.
  }
  \label{tab:mnist_map}
\end{table*}

\begin{figure} [t]
  \centering
  \includegraphics[angle=0, height=0.25\textwidth]{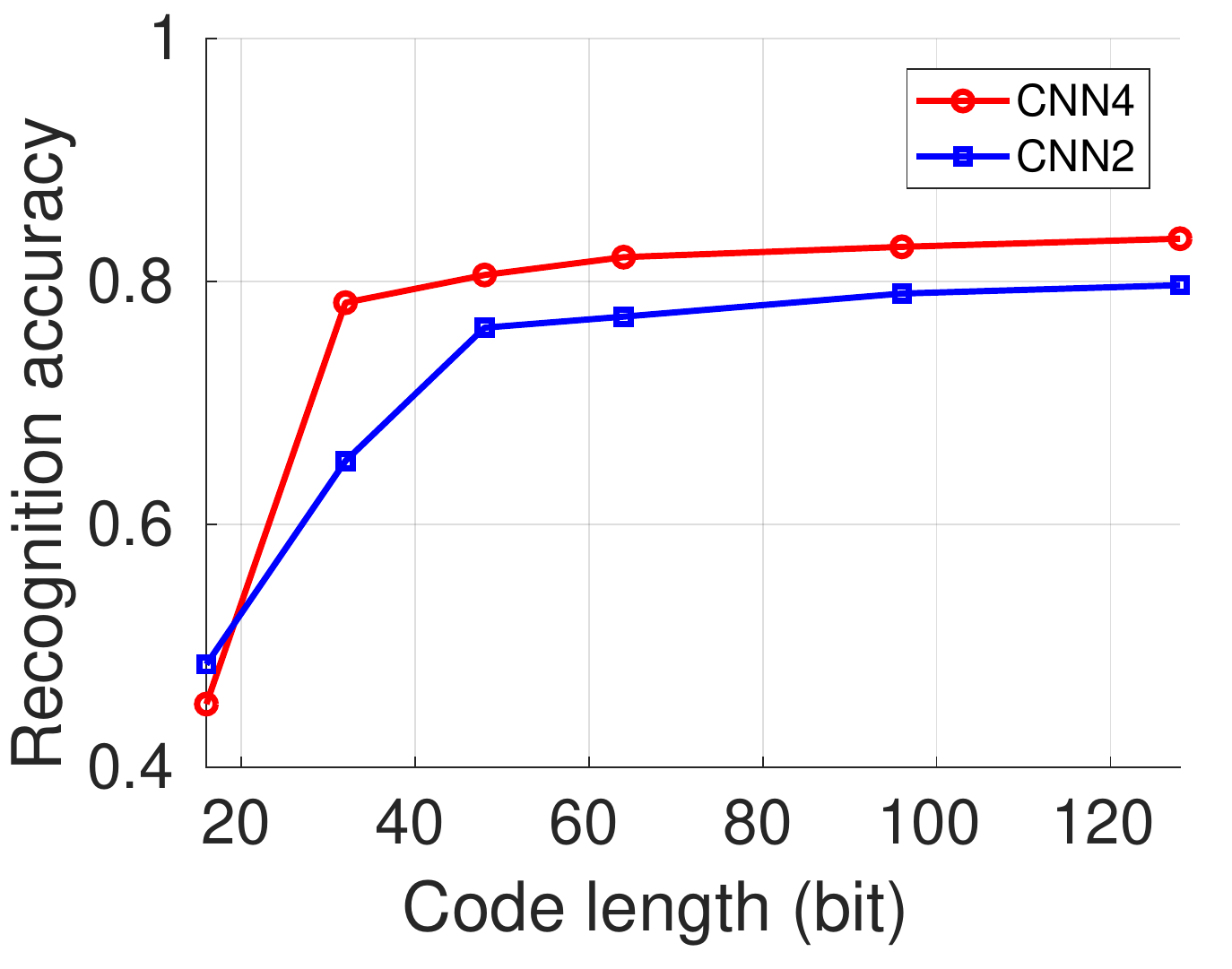}
  \caption{The forest boosting effect using ForestHash codes.  ForestHash shows
    a principled and robust procedure to train and deploy in parallel an
    ensemble of light-weight CNNs, instead of simply going deeper. }
  \label{fig:forestboost}
\end{figure}


\begin{table*}[ht]
  \centering

  {\scriptsize
    \begin{tabular}{l|l|l|l|l|l|l|l}
      \hline
      ForestHash  & ForestHash & CM-SSH \cite{CM-SSH}&  CM \cite{wikixmod} & SM \cite{wikixmod}& SCM\cite{wikixmod}& MM-NN\cite{MM-NN}& CM-NN \cite{MM-NN}\\
      (64-bit) & (36-bit) &&&&&& \\
      \hline
      \hline
      \textbf{50.8} & 45.5 & 18.4 & 19.6 & 22.3 & 22.6 & 27.4 & 25.8 \\
      \hline
    \end{tabular}
  }
  \caption{Cross-modality image retrieval using text queries on the Wiki dataset
    (mAP in \%).}
  \label{tab:wiki}
\end{table*}

\subsection{Image retrieval}
We first adopt a CIFAR-10 protocol popular among many deep-learning based
hashing methods, e.g., DSH \cite{DSH}, where the official CIFAR-10 train/test
split is used; namely, 50,000 images are used as the training and the gallery,
and 10,000 images as the query.  Table~\ref{tab:CIFAR-hash-50k} reports the
retrieval performance comparisons with multiple hashing methods\footnote{Results
  are taken from the respective papers.}. ForestHash with a simplest two-layer
learner {CNN2} in Table~\ref{tab:cnn} already significantly outperforms
state-of-the-art methods. Given such large size of training set, retrieval
performance increases using more complex network structures as learners, e.g.,
{CNN4} or {VGG16} over {CNN2}.

The superior retrieval performance of the ForestHash codes in
Table~\ref{tab:CIFAR-hash-50k} can be easily explained by both the low-rank loss
properties in Figure~\ref{fig:cifar_tsne} and the boosting effect of the random
forest in Figure~\ref{fig:forestboost}.  ForestHash shows a principled and
robust procedure to train and deploy in parallel an ensemble of light-weight
CNNs, instead of simply going deeper.  As shown in Table~\ref{tab:CIFAR-acc},
ForestHash performs at the level of other state-of-the-art image classification
techniques, e.g., ResNet, while utilizing a 128-bit only representation.

We further experiment with CIFAR-10 using reduced size of training with both handcrafted feature and deep features.  We adopt
the same setup as in \cite{KSH, sparsehash} for the image retrieval experiments:
we only used 200 images from each class for training; and for testing, a
disjoint test set of 1,000 images are evenly sampled from ten classes, to query
the remaining 59,000 images.  Images are used as inputs for ForestHash with CNN
learners, and 384-dimensional GIST descriptors are used for other compared
methods, including ForestHash with an RBF kernel.

Table~\ref{tab:cifar10} summarizes the retrieval performance of various methods
on CIFAR-10 at reduced training using the mean precision and recall for Hamming
radius 0 and 2 hash look-up.  For the compared methods SH, KSH, AGH1 and AGH2, we
use the code provided by the authors; while for SparseHash, we reproduce the
results from \cite{sparsehash}. SH is unsupervised, while the rest of the
hashing schemes are all supervised.  We report the performance of ForestHash
using the random, unsupervised, supervised, and semi-supervised hash aggregation
schemes, respectively.  We observe that the proposed information-theoretic code
aggregation provides an effective way to combine hashes from different trees,
and showing further benefits to unify both unsupervised and supervised
aggregation.  We also observe that using softmax loss only for CNN learners
leads to performance degradation.
 At reduced training, more complex learner
structures show no obvious advantage.  In general, the proposed ForestHash shows
significantly higher precision and recall compared to other methods.

The supervised hashing methods HDML \cite{HDML}, TSH \cite{tsh}, and FastHash
\cite{fasthash} report excellent performance, where HDML is a deep learning
based hashing method, and FastHash is a boosted trees based method.  We adopt
the experimental setting from \cite{HDML}, i.e., a 60K training set and a
disjoint 10K query set split on the MNIST data.  Each hashing method is assessed
by the retrieval precision and recall at radius 0.  As shown in
Table~\ref{tab:mnist}, using all 60K training samples, ForestHash with an RBF
kernel shows comparable performance as HDML and FastHash, and better than
TSH. ForestHash with a two-layer CNN significantly outperform all compared
methods.  We further assume labels are only available for a small subset of
data, which is often the case for a retrieval system. When the number of labeled
samples reduces to 100 and 30 per class respectively (instead of 6,000), the
retrieval performance of other deep learning and boosted tree-based hashing
degrades dramatically, as those methods require a dense training set to learn a
rich set of parameters.  Due to the subspace assumption behind the low-rank
loss, which are known to be robust in the regime with few labeled training
examples per class \cite{revlearn}, ForestHash significantly outperforms
state-of-the-art methods for such reduced training cases.  Note that the
training and hashing time of ForestHash reported here is the time for one tree,
in order to emphasize the fact that different trees are trained and deployed
independently and can easily be done in parallel

More experiments were conducted on MNIST following \cite{AGH}, enabling the
comparison with more hashing methods for which we have no implementation
accessible. We split the MNIST dataset into a training set containing 69,000
samples and a disjoint query set of 1,000 samples. Table~\ref{tab:mnist_map}
reports the Hamming ranking performance measured by the Mean Average Precision
(mAP) (performance of other methods is reproduced from \cite{AGH}).  For both
code lengths, the proposed ForestHash significantly outperforms other hashing
methods.

\subsection{Cross-modality retrieval}

\begin{figure} [t]
\centering
 \includegraphics[angle=0, height=0.34\textwidth]{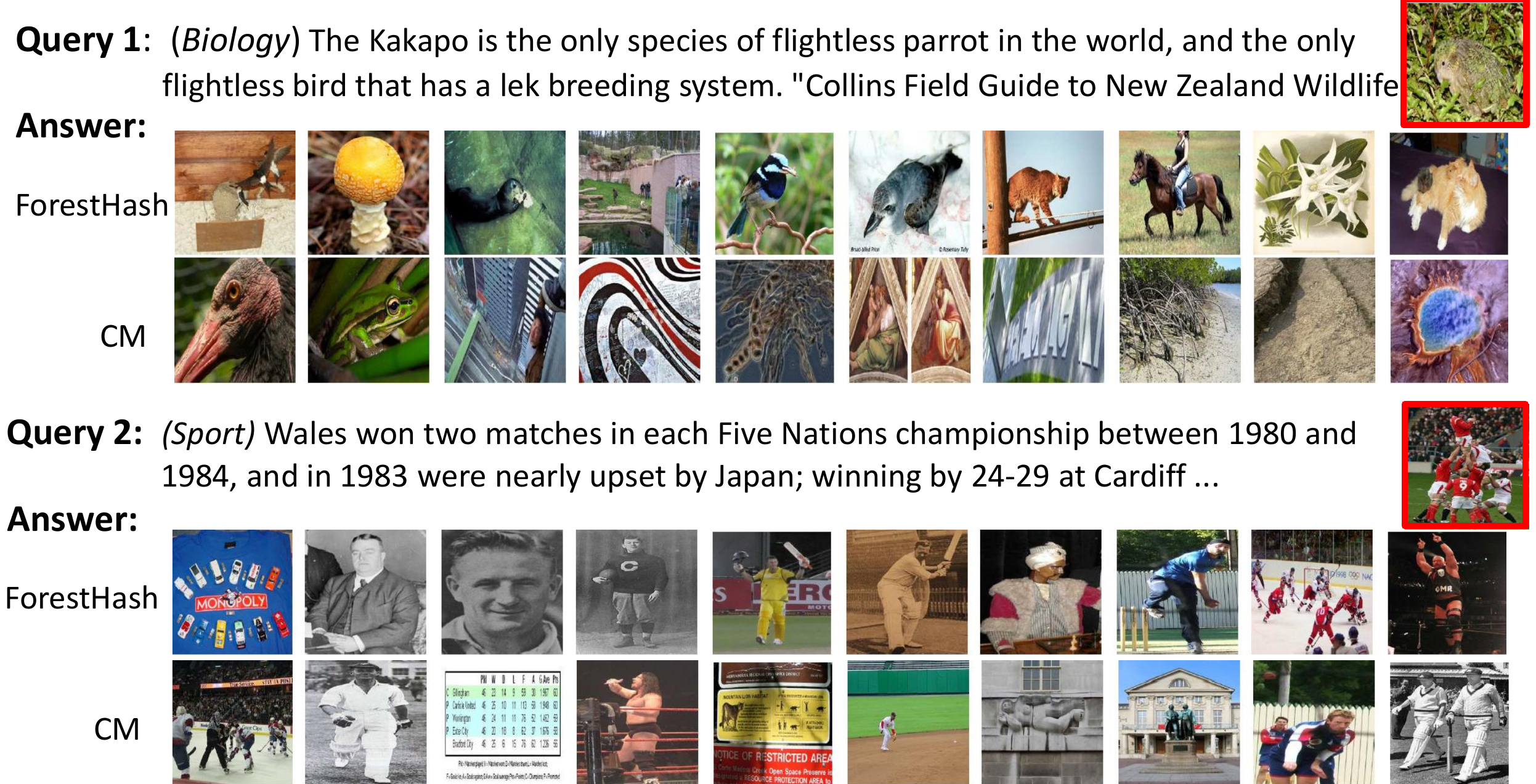}
\caption{Two examples of text queries and the top-10 images retrieved by
  ForestHash and CM \cite{wikixmod}.  Note that only text information are used
  to compose each query, and images are retrieved from the same category of the
  query text.  }
\label{fig:wiki}
\end{figure}

We performed a cross-modality retrieval experiment following \cite{MM-NN,
  wikixmod} on the Wikipedia dataset.  The purpose is to demonstrate that
ForestHash natively supports cross-modality, though not being designed for.  The Wikipedia dataset contains a
total of 2866 documents. These are article-image pairs, annotated with a label
from 10 semantic classes.  To enable a fair comparison, we adopted the provided
features for both images and text from \cite{wikixmod}.  Table \ref{tab:wiki}
shows the mean average precision scores for the cross-modality image retrieval
using text queries.  The proposed ForestHash significantly outperforms
state-of-the-art hashing approaches on cross-modality multimedia retrieval
tasks. Note that MM-NN and CM-NN \cite{MM-NN} in Table \ref{tab:wiki} are both
deep learning motivated hashing methods.  Two examples of cross-modality text
queries and the top-10 images retrieved are shown in Figure~\ref{fig:wiki},
using ForestHash and CM \cite{wikixmod}.  Note that only text information is
used to compose a query, and ForestHash retrieves images from the same category
of the query text. ForestHash significantly outperforms CM with codes at least
$10 \times$ shorter.

\section{Conclusion} \label{sec:con}
Considering the importance of compact and computationally efficient codes, we
introduced a random forest semantic hashing scheme, 
 extending random forest  beyond classification
and for large-scale multimodal retrieval of incommensurable data.  The proposed
scheme consists of a forest with random class grouping, low-rank loss,  and an
information-theoretic code aggregation scheme.  Using the matrix nuclear norm as
the optimization criterion, the low-rank loss simultaneously reduces variations
within the classes and increases separations between the classes.  Thus, hash
consistency (similarity) among similar samples is enforced in a random tree.
The information-theoretic code aggregation scheme provides a nearly optimal way
to combine hashes generated from different trees, producing a unique code for
each sample category, and is applicable in training regimes ranging from totally
unsupervised to fully supervised.  Note that the proposed framework combines in
a fundamental fashion  kernel methods, random forests, CNNs, and
 hashing.  Our method shows exceptional effectiveness in
preserving similarity in hashes, and significantly outperforms state-of-the-art
hashing methods in large-scale single- and multi-modal retrieval tasks.

\section*{Acknowledgements}
Work partially supported by AFOSR, ARO, NGA, NSF, ONR. Jos\'e Lezama was supported by ANII (Uruguay) grant PD\_NAC\_2015\_1\_108550.

\bibliographystyle{splncs}
\bibliography{hash_eccv}

\end{document}